# Deep Learning in the Automotive Industry: Recent Advances and Application Examples

**Kanwar Bharat Singh and Mustafa Ali Arat**

Tire Intelligence - Innovation Technology, Goodyear Innovation Center, Luxembourg
Corresponding author: Kanwar B. Singh (kanwar-bharat_singh@goodyear.com)

**ABSTRACT** One of the most exciting technology breakthroughs in the last few years has been the rise of deep learning. State-of-the-art deep learning models are being widely deployed in academia and industry, across a variety of areas, from image analysis to natural language processing. These models have grown from fledgling research subjects to mature techniques in real-world use. The increasing scale of data, computational power and the associated algorithmic innovations are the main drivers for the progress we see in this field. These developments also have a huge potential for the automotive industry and therefore the interest in deep learning-based technology is growing. A lot of the product innovations, such as self-driving cars, parking and lane-change assist or safety functions, such as autonomous emergency braking, are powered by deep learning algorithms. Deep learning is poised to offer gains in performance and functionality for most ADAS (Advanced Driver Assistance System) solutions. Virtual sensing for vehicle dynamics application, vehicle inspection/heath monitoring, automated driving and data-driven product development are key areas that are expected to get the most attention. This article provides an overview of the recent advances and some associated challenges in deep learning techniques in the context of automotive applications.

## I. INTRODUCTION

From voice assistants to self-driving cars, deep learning (DL) is redefining the way we interact with machines. DL has been able to achieve breakthroughs in historically difficult areas of machine learning such as text-to-speech conversion, image classification, and speech recognition. DL generally refers to a class of models and algorithms based on deep artificial neural networks (ANN). Perceptron, the building block of an ANN was first presented in the 1950s (Figure 1). ANNs were a major area of research in both neuroscience and computer science until the late 1960s and the technique then enjoyed a resurgence in the mid-1980s. At Bell Labs, Yann LeCun developed a number of DL algorithms [1] in the late 1980s, including the convolutional neural network (CNN). Pioneering deep neural networks by Yann LeCun could classify handwritten digits [2] with good speed and accuracy and were widely deployed to read over 10% of all the cheques in the United States in the late 1990s and early 2000s.

Even though the basic theories related to ANN have existed since the 1950s, the field of DL has matured a lot in the last decade and changed a lot in the last few years. The "deep" in deep learning is not a reference to any kind of deeper understanding achieved by the approach; rather, it stands for the many layers in the neural network that contribute to a model of the data [4]. There are four key factors that are driving the progress and uptake of DL.

1. More Compute Power: e.g. graphics processing unit (GPUs), tensor processing unit (TPUs) [5] etc.
2. Organized Large Datasets: e.g. ImageNet [6] etc.

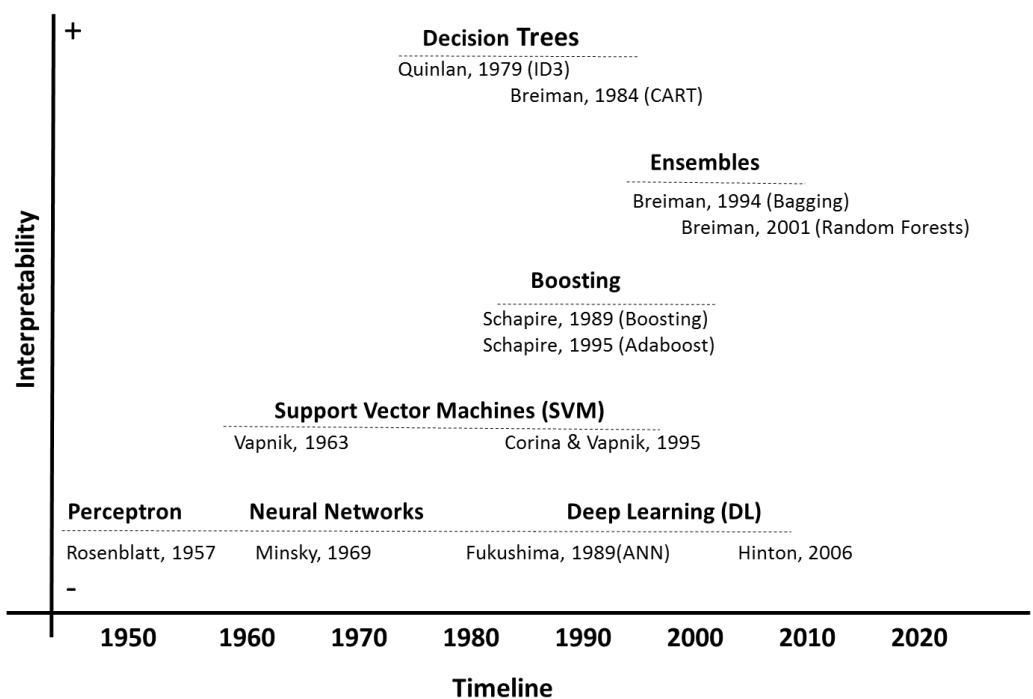

**FIGURE 1.** Evolution of learning algorithms [3].

3. Better Algorithms: e.g. activation functions like ReLU [7], optimization schemes etc.
4. Software & Infrastructure: e.g. Git, Robot Operating System (ROS) [8], TensorFlow [9], Keras [10] etc.

Among these four factors, increased compute power has been the main workhorse that is making DL a burgeoning field of artificial intelligence (AI). New architectures (Figure 2) are scaled to be deeper, taking advantage of much larger datasets and parallel computing power. This newly gained capability that increased compute power has brought is now laying the foundation for DL to disrupt numerous industries. The increased proliferation of data being produced and stored made the healthcare industry an early adopter of DL technology. New DL-based tools are shaping the future of patient care. Future applications of DL in the healthcare field are expected to extend beyond imaging data to include electronic health records [11] etc.

More recently, DL is also being employed to solve real-world problems that are relevant for the automotive industry. Historically, the automotive industry has been a generator of large volumes of data. Data are collected from suppliers during the design, development and testing of various subsystems, real-time data are generated by hundreds of on-vehicle sensors, system faults are recorded by the on-board diagnostics (OBD), and information is collected about the vehicle servicing history and customer preferences at dealerships/service centers. Going forward, increasingly connected and autonomous vehicles will generate even larger amounts of data. For instance, some estimates suggest that a single autonomous vehicle will generate 4,000 GB of data each day [12] (Figure 3).

In the past, it was difficult to mine and analyze these data in an efficient, fast, and automated manner, thus leaving vast amounts of valuable information untapped and underutilized. This is set to change as opportunities for connectivity exponentially improve, DL matures, and researchers explore new use cases for the technology in the automotive industry. An increasing appetite from customers for greater connectivity, intelligence and safety in their vehicles is expected to drive the adoption of DL as a mainstream technology in the automotive industry.

The purpose of this article is to provide a timely review and an introduction to DL algorithms applied in the context of automotive applications. It is aimed to provide readers with a background on different DL architectures and also the latest developments as well as achievements in this area. The rest of the paper is organized as follows. In Section 2, the main DL architectures are reviewed. The usage of these DL architectures in vehicular applications are highlighted in Section 3. Conclusions and future topics of research are presented in Section 4.

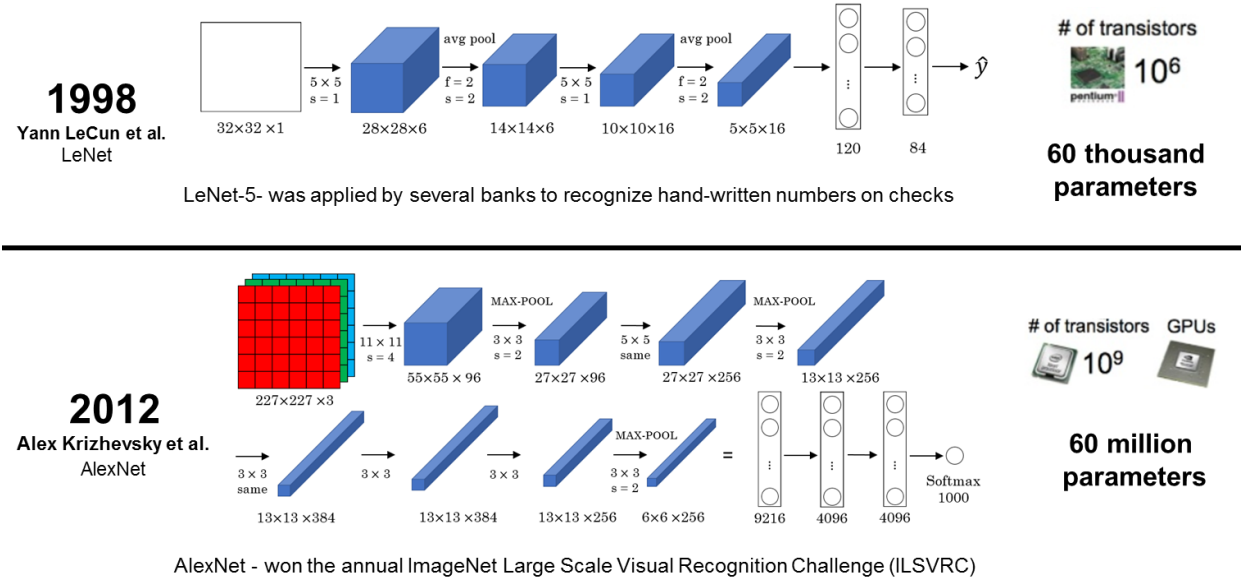

**FIGURE 2.** Pioneering work, LeNet versus a modern incarnation, AlexNet [2], [13], [14].

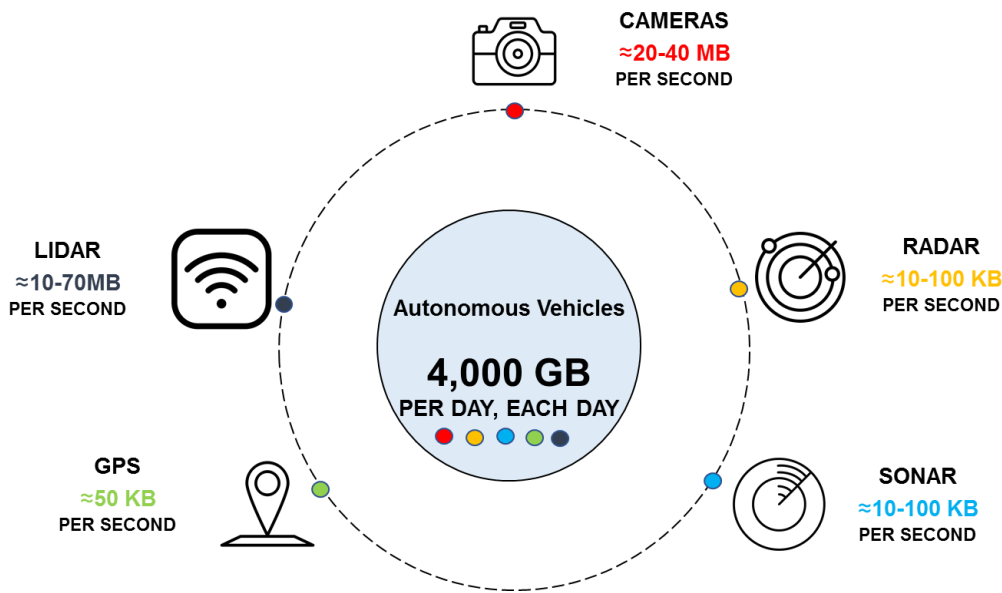

**FIGURE 3.** Estimates of data generated from an autonomous vehicle [12].

## II. ARCHITECTURE OF DEEP LEARNING MODELS

Conventional machine learning models are trained on features extracted from the raw data using feature extraction and feature engineering techniques. This feature vector is used by the machine learning system, often a classifier, to detect or classify patterns in the input. Designing a feature extractor that can successfully transform raw data into a suitable feature vector requires considerable domain expertise. Moreover, hand-engineered features are time consuming, brittle and not scalable in practice. This is especially true in the case of unstructured data. DL solves the representation problem for unstructured data, which is why it performs so much better than other algorithms for unstructured data. The key differentiating point of DL is that the model learns useful representations and features automatically, directly from the raw data, bypassing this manual and difficult step of hand engineering the feature vector.

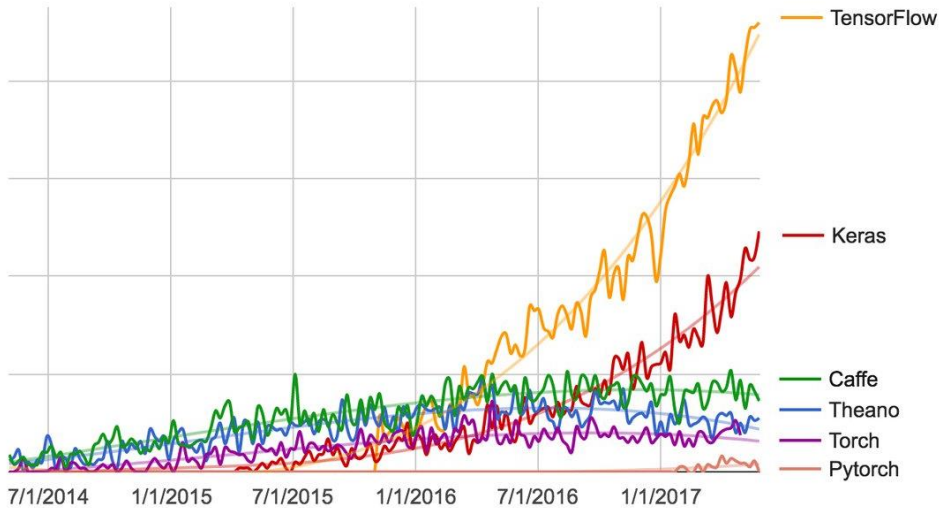

**FIGURE 4.** Search interest for the different deep learning frameworks [4].

TABLE I
COMPARISON OF MODEL TRAINING TIMES [15]

| Platform | Time |
|---|---|
| Laptop PC | 21.6 hr. |
| Cloud (4x CPUs) | 16.1 hr. |
| Cloud (16 CPUs) | 4.8 hr. |
| Cloud (1x GPU) | 4.72 hr. |
| Cloud (2x GPUs) | 2.5 hr. |
| Cloud (4x GPUs) | 1.6 hr. |

Most commonly used libraries for training DL models include:
- TensorFlow [9], a framework originating from Google Research
- Keras [10], a DL library originally built by Francois Chollet and recently incorporated in TensorFlow
- Caffe [16], a DL framework, originally developed at University of California, Berkeley
- Theano [17], a Python library and optimizing compiler primarily developed by the Montreal Institute for Learning Algorithms (MILA) at the Université de Montréal
- Microsoft Cognitive Toolkit, previously known as CNTK [18]
- Pytorch [19], a framework associated with Facebook Research

Apart from these libraries, there are other open source DL software frameworks available such as Apache MXNet and Gluon. All these libraries are open source and under active development. Over the years, TensorFlow has emerged as the most popular framework (Figure 4).

Training a DL algorithm is a computationally intensive process. It involves repetitive arithmetic operations, requiring multi-core computing power due to the limited number of arithmetic logic units (ALUs) on CPUs. On the other hand, graphical processing units (GPUs) are inherently designed for repetitive arithmetic operations. They have thousands of ALUs and are naturally more suited for DL applications. Modern GPUs contain numerous simple processors (cores) and are highly parallel, which makes them also very effective in running/embedding some algorithms. Matrix multiplications, the core of DL right now, are among these. As demonstrated by the results in Table I, it takes the power of 16 CPUs to match the power of 1 GPU [15].

Recent DL models have numerous hidden layers, hundreds of millions of weights, and billions of connections between units.

TABLE II
LIST OF CHIPS FOR IMPLEMENTING REAL-TIME DEEP LEARNING APPLICATION

| Manufacturer | Chip | Source |
|---|---|---|
| NVIDIA | Volta | https://www.nvidia.com/en-gb/data-center/volta-gpu-architecture/ |
| Mobileye | EyeQ | https://www.mobileye.com/our-technology/evolution-eyeq-chip/ |
| Intel | Movidius | https://www.movidius.com/ |
| Qualcomm | Vision Intelligence 300/400 Platform | https://www.qualcomm.com/media/documents/files/qualcomm-vision-intelligence-300-400-platforms.pdf |
| Samsung | Exynos 9 Series 9810 | https://www.samsung.com/semiconductor/insights/tech-leadership/the-exynos-9-series-9810-processor-redefines-boundaries/ |
| Apple | A11 Bionic | https://de.wikipedia.org/wiki/Apple_A11_Bionic |
| Google | Tensor Processing Unit | https://en.wikipedia.org/wiki/Tensor_processing_unit |

Whereas training such large networks could have taken weeks only a few years ago, progress in hardware, software and algorithm parallelization have reduced training times to a few hours. To be able to optimally utilize the computing power of the hardware (GPU or CPU) in use, the above-mentioned libraries rely on so-called support libraries.

For the GPUs, NVIDIA introduced CUDA and cuDNN, which allow the use of popular languages such as C, C++, Fortran, Python and MATLAB and express parallelism through extensions in the form of a few basic keywords. For CPUs, Intel introduced the Math Kernel Library (MKL) which is optimized to speed up arithmetic operations.

A number of companies such as NVIDIA, Mobileye, Intel, Qualcomm, Apple and Samsung (Table II) are also developing specialized chips to enable embedded implementation of real-time DL applications in smartphones, cameras, robots and self-driving cars.

The most common models in DL are types or subclasses of ANN with architectures specifically designed for their respective application niches. The two most commonly used variants are the convolutional neural networks (CNN) and the recurrent neural networks (RNN). CNNs have brought about breakthroughs in processing images, video, speech and audio, whereas RNNs have excelled in dealing with problems involving sequential data such as text and speech [20].

*A. CONVOLUTIONAL NEURAL NETWORKS (CNN)*
CNN-based DL models for image classification have achieved an exponential decline in error rate through the last few years and have surpassed the level of human accuracy [21]. A typical CNN is composed of numerous layers. CNN uses convolutions within the node operations instead of vanilla matrix multiplication operations as in the case of classic ANNs. Training a CNN requires a lot of computing power, labelled images (typically several thousand images for each object class) and weeks of processing time. Most applications suffer from a deficit of training image samples, and therefore training a CNN from scratch becomes impractical. To overcome this limitation, researchers use a technique called "transfer learning" [22]. The concept of transfer learning refers to taking a pre-trained network and modifying it for your own problem. The idea of transfer learning is inspired by the fact that researchers can intelligently apply knowledge learned previously to solve new problems. Table III contains a short list of some famous CNN architectures.

*B. RECURRENT NEURAL NETWORKS (RNN)*
Many applications have temporal dependencies, i.e. the current output is not only a function of the current input but also the previous inputs. RNNs are neural networks that can capture temporal dependencies and hence are well suited for processing sequence data for making predictions. They process sequence or temporal data by iterating through the sequence elements and maintaining a state containing information relative to what it has seen so far (Figure 5).

One major issue with simple RNNs (with vanilla architecture) [23] is that as the number of time steps increases, it fails to derive context from time steps which are much far behind. To understand the context at time step t+1, we might need to know the representations from time steps 0 and 1. But, since they are so far behind, their learned representations cannot travel far ahead to influence at time step t+1. This is due to the vanishing gradient problem, an effect that is similar to what is observed with non-recurrent networks (feedforward networks) that are many layers deep: as you keep adding layers to a network, the network eventually becomes untrainable [4]. The vanishing gradient problem is an issue that arises when training algorithms through gradient descent. Gradients control how much the network learns during training, if the gradients are very small or zero, then little to no training can take place, leading to poor predictive performance The theoretical reasons for this effect were studied by Hochreiter, Schmidhuber, and Bengio in the early 1990s [28]. Two popular RNN architectures have been designed to overcome this problem (Figure 6). They are better known in literature as the Long Short Term Memory (LSTM) [29] and Gated Recurrent Units (GRU) [30] (Figure 6)

TABLE III
POPULAR CNN ARCHITECTURES

| CNN Model | Year | No of convolution layers | No of parameters (millions) | Reference |
|---|---|---|---|---|
| AlexNet | 2012 | 5 | 62.4 | [13] |
| VGG-16 Net | 2013 | 16 | 138.3 | [24] |
| ResNet-50 | 2015 | 50 | 25.6 | [25] |
| GoogLeNet | 2015 | 22 | 7 | [26] |
| SqueezeNet | 2016 | 18 | 1.2 | [27] |

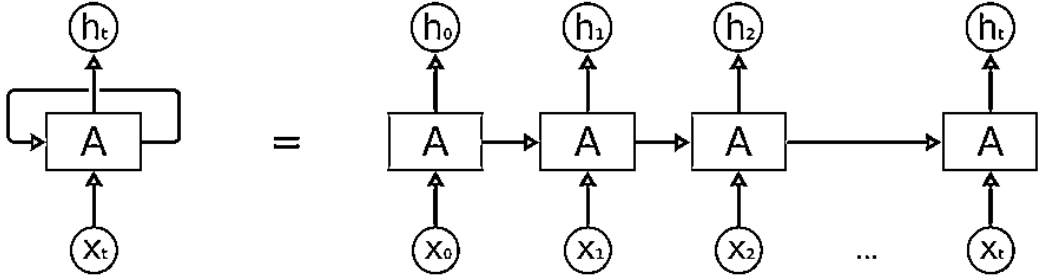

FIGURE 5. Structure of RNNs.

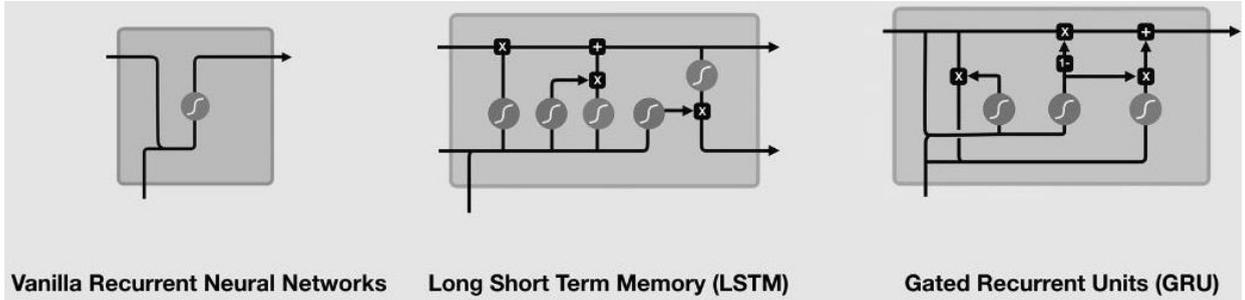

FIGURE 6. RNN architectures.

## III. DEEP LEARNING BASED AUTOMOTIVE APPLICATIONS: NOVEL USE CASES

This section summarizes novel use-cases that utilize DL principles for enabling automotive related applications. The selected use-cases are grouped under four key topics.
1. Virtual sensing for vehicle dynamics
2. Vehicle inspection/health monitoring
3. Automated driving
4. Data-driven product development

The main rationale behind selecting these topics is their relevance for the automotive industry.

The first topic focuses on the so-called soft/virtual sensors that can be used for estimating relevant vehicle dynamic and tire-related states. It is well documented in literature [31] that certain states of the vehicle motion and environmental conditions are crucial for safely negotiating a maneuver, e.g. vehicle body sideslip and/or tire-road friction coefficient are crucial for the implementation of an advanced vehicle stability control system [32], [33]. Although physical sensors exist to measure these states of interest, they are designed for laboratory use and excessively cost exorbitant for usage in mass production vehicles. For instance, an automotive-grade optical sensor for measuring the vehicle sideslip angle can cost upwards of $30,000, which renders it impractical for use in a production vehicle. Researchers generally resort to state estimators and signal processing methods [31], [34] for computing these states from already existing measurements. Nevertheless, designing and tuning such systems are also

considerably costly and rely on linearized models or assumptions regarding the vehicle motion which do not hold valid at all times. This typically results in large variation in their accuracy and renders them as unusable for many applications. DL techniques offer the promise of developing solutions that can be immune to such problems.

The second topic is on inspection and health monitoring for vehicle parts. As a vehicle rolls out of the production line, it has expected expiration dates for different parts, yet many exogenous factors during a daily drive change these expectations. There are sensors for warning the driver of such cases; however, such sensors are primarily used on more critical parts of the vehicle that might prevent a mechanical failure/breakdown. DL-enabled algorithms provide opportunities to estimate these problems (e.g. mechanical failure, aging, etc.), as well as to detect and warn about failing parts.

The third and by far the largest topic is on the applications of DL in autonomous driving. This part, as of the date of preparation of this study, covers the most popular and promising application subjects. As the literature is already vast with many valuable publications and patents, the section summarizes select ones as the most striking examples. The most amount of studies seem to be shown in the visual applications and image processing for object identification that are essentially used in identifying road users, traffic signs, lanes and other objects. These applications are followed by implementations in route planning, as either the so-called end-to-end solutions (i.e. actuators commands are computed by the network) or as location and situation awareness systems (i.e. providing a complete environmental model for the actuators).

The fourth topic is on the usage of data and advanced algorithms in the product design and development.

*A. VIRTUAL SENSING FOR VEHICLE DYNAMICS*

Effective implementation of automotive stability control systems largely depends on accurate vehicle dynamic state information. Therefore, knowledge of certain tire-vehicle dynamic states is of immense importance for vehicle motion and control. Vehicle sideslip angle (β) and the tire-road friction coefficient (μ) are probably the two most prominent vehicle states of interest.

Vehicle sideslip angle is a strong indicator of the vehicle's stability. Hence, there is a great level of interest in using DL techniques for estimating the vehicle sideslip angle (Table IV). The authors from Bosch Engineering and Politecnico di Torino [35] investigated the use of an RNN with a LSTM cell. They used conventional sensor data sampled at 500 Hz, namely lateral and longitudinal acceleration, yaw-rate, the front and rear steering angle, the vehicle speed and the wheel speeds for the four wheels. Thereafter, they used an 8-layer LSTM network with neurons varying from 40 to 100 in number. As the readers would also possibly appreciate, the authors utilized an extensively large network which would present the pitfall for overtraining. The authors did not explicitly state the size of their dataset; however, they explain it was generated during an extensive test campaign consisting of standard international standard organization (ISO) test maneuvers and also high speed circuit laps. Results presented in this work are promising regarding the robustness of the method. On the other hand, the study does not mention about robustness of the proposed approach to varying environmental conditions such as tire-road surface conditions.

A second study conducted by researchers from Porsche AG and University of Stuttgart [36] follows a similar approach to generate and train an RNN this time with the GRU cell. In addition, they propose combining the inputs to the network with the outputs of a kinematic model which computes the rate of change of the vehicle sideslip angle in time. Their dataset sampled at a 100 Hz consisted of nearly 6 million datapoints on dry, wet, and snowy road surface conditions. As a result, they were able to provide a more extensive comparison on model sensitivity.

The above two studies conclude that DL methods for vehicle sideslip estimation result in a significant improvement over classical state estimation techniques using model-based and/or kinematic-based observers [37]–[39]. Nevertheless, authors of both these studies agree that the structure of the network (i.e. hyper parameters) needs to be carefully optimized which to the best of the authors' knowledge has not been analyzed in any study yet and hence could be a potential area for future research.

Apart from vehicle sideslip angle, the other key parameter of interest is the grip level between the tire and the road. The importance of road-friction estimation is reflected by the considerable amount of work that has been done in this field [40]–[43]. Lately, we see a surge in the usage DL models for road-friction estimation. In [44], authors used LSTM-RNNs for detecting road wetness from audio of the tire-surface interaction and discriminating between wet and dry classes. Although the authors claim an outstanding performance on the road wetness detection task with an 93.2% unweighted average recall (UAR) for all vehicle speeds, it is noteworthy to highlight the fact that the model was trained on considerably limited dataset (one vehicle, limited routes, limited driving time etc.). Moreover, the authors haven't considered the fact that tire-road noise is heavily impacted by the tire's structural properties (e.g. tread pattern, pitch sequence etc.), and also the tire wear condition. Hence, the robustness of the proposed approach is contentious.

What is emerging as a more promising approach for road-friction estimation is the usage of vision-based DL models [45]–[47]. This is primarily due to the availability of pre-trained CNN models that have demonstrated reliable prediction performance on a heterogeneously large datasets such as the ImageNet [6]. The authors of [45]–[47] utilize the transfer learning methodology [22] to build image classification models, wherein the different classes correspond to the road surface conditions (e.g. dry, wet, snow, ice). In [45], the author trains a binary image classification model with images labelled as high friction if μ > 0.6 and medium friction if 0.2< μ < 0.6. Moreover, he also applied data augmentation techniques to effectively increase the size of the data set, thus reducing the risk of overfitting. A DenseNet [48] model was shown to achieve the highest prediction accuracy at just above 90 %. Researchers from Volvo Cars Technology [46] applied the SqueezeNet [27] model on images extracted from public domain YouTube videos. The model showed a 94-99% classification accuracy for dry, wet/water, slush and snow/ice conditions. None of these papers [45]–[47] have evaluated the impact of variations in the camera image quality as a results exogenous noise coming from vibrations, unfavorable lighting conditions, inclement weather impacts etc. These issues will have to be addressed before we can see these models finding their way into production vehicles.

TABLE IV
SUMMARY OF TECHNIQUES FOR VIRTUAL SENSING FOR VEHICLE DYNAMICS

| Application Enabled | Vehicle Sideslip Estimation | | Road Friction Estimation | | | |
|---|---|---|---|---|---|---|
| Data Type | Time series (500 Hz) | Time series (100 Hz) | Time series | Images | Images | Images |
| Data Source | Proprietary | Proprietary | Proprietary | Proprietary | Public | Public and Proprietary |
| Data Size | Not specified objectively | ~16 hours of driving | ~2 hours of driving | 37,000 images | 5,260 images | 15,000 images |
| Deep Learning Algorithm | RNN - LSTM | RNN - GRU | RNN - LSTM | CNN | CNN | CNN |
| Platform Used for Model Learning | GPU | Not specified | Not specified | GPU | Not specified | GPU |
| Accuracy Achieved | Not specified objectively | Dry 99.43% Wet 97.86% Snow 78.95% | 93.2% (recall) | 90.02% | 97% | 84% |
| Reference | [35] | [36] | [44] | [45] | [46] | [47] |

*B. VEHICLE INSPECTION/HEALTH MONITORING*
Vehicle repair workshops and insurance companies are beginning to tap into the potential of DL [49], [50]. DL technology is being used to automate visual tasks, such as inspecting the damage to a vehicle. This approach provides drivers and insurance companies with crucial information about the extent of the damage and also presents tremendous opportunities to reduce the cost of manual labor. For instance, UVeye's 360° system scans and automatically detects damages in the outer frame of a vehicle [51]. Vision based algorithms are also being proposed as an effective approach in preventing fraud claims as well as meeting the requirement to speed up the insurance claim pre-processing [52].

With the advent of car sharing fleets, the automotive market is slowly but steadily transitioning from an individual ownership model to a fleet ownership model. With all the interest around DL, it is natural for car-sharing fleets to explore the options regarding predictive and prognostic maintenance of vehicles [53], [54]. We are already starting to see commercial offerings from OEMs that combine data from sensors and advanced algorithms to monitor the vehicle battery, starter, and fuel pump each time the car is started [55]. Advanced predictive maintenance enabled by DL techniques is one of the most widely accepted benefits of the fourth industrial revolution. With predictive maintenance, faults can be reliably detected before they occur based on data anomalies. A good example is in [56] where the authors use a CNN for detecting defects in a tire. In another study [57], authors demonstrate the usage of CNNs for tire leakage detection. There are numerous studies focused on diagnostic system for electric vehicle [58].

### C. AUTOMATED DRIVING

Most established OEMs as well as upcoming startups and mobility service providers like Uber, Cruise and Waymo are working on building autonomous vehicles (AVs). These AVs use sensors like camera, radar, LiDAR, etc. (Figure 7), generating massive amounts of data every second.

Unlike traditional vehicles that mostly employed a model-based approach with hand-coded decision logic for vehicle control, most AVs typically leverage an end-to-end DL framework for vehicle control. In the case of an end-to-end framework, the entire decision-making logic is encoded into a neural network space that produces the driving decisions (Figure 8). A few interesting examples include the work done by researchers from NVIDIA Corporation [59] showing how CNNs are able to learn the entire task of lane and road following without manual decomposition into road or lane marking detection, semantic abstraction, path planning, and control. Similarly, researchers from the University of California, Berkeley [60] proposed an approach to learn a generic driving model from large scale crowd-sourced video dataset with an end-to-end trainable architecture

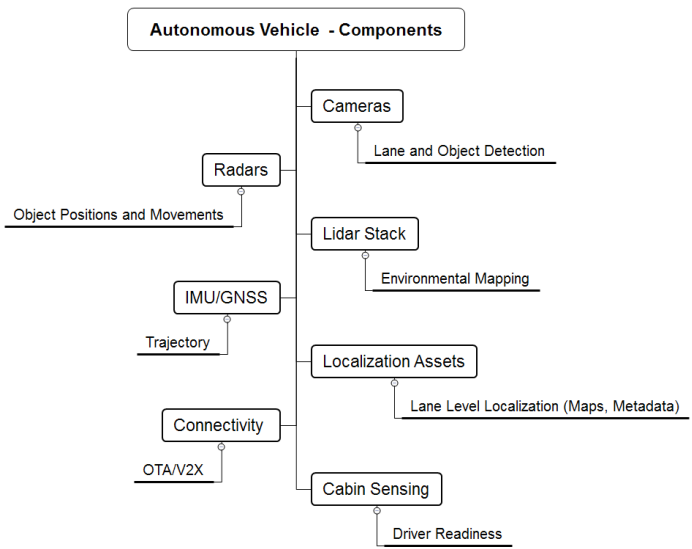

**FIGURE 7.** Autonomous vehicle componentry and their intended function.

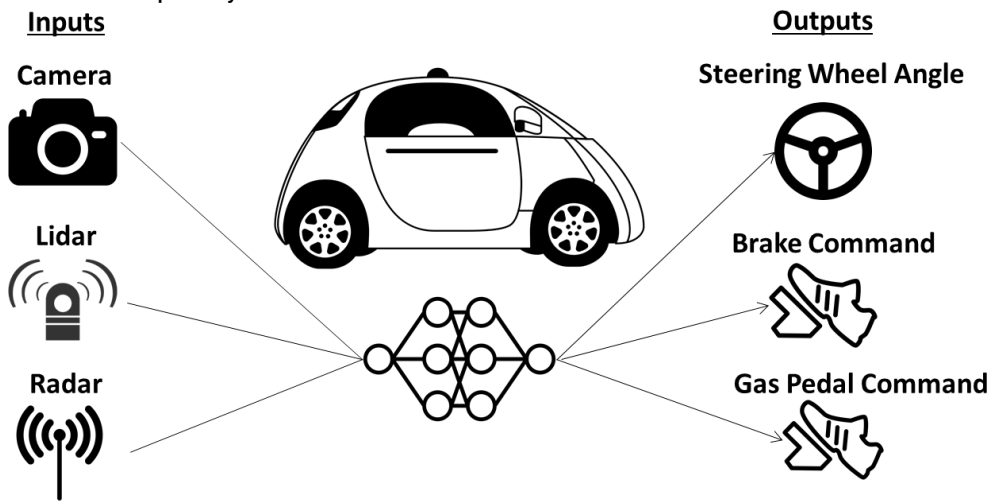

**FIGURE 8.** End-to-end deep learning framework – illustration.

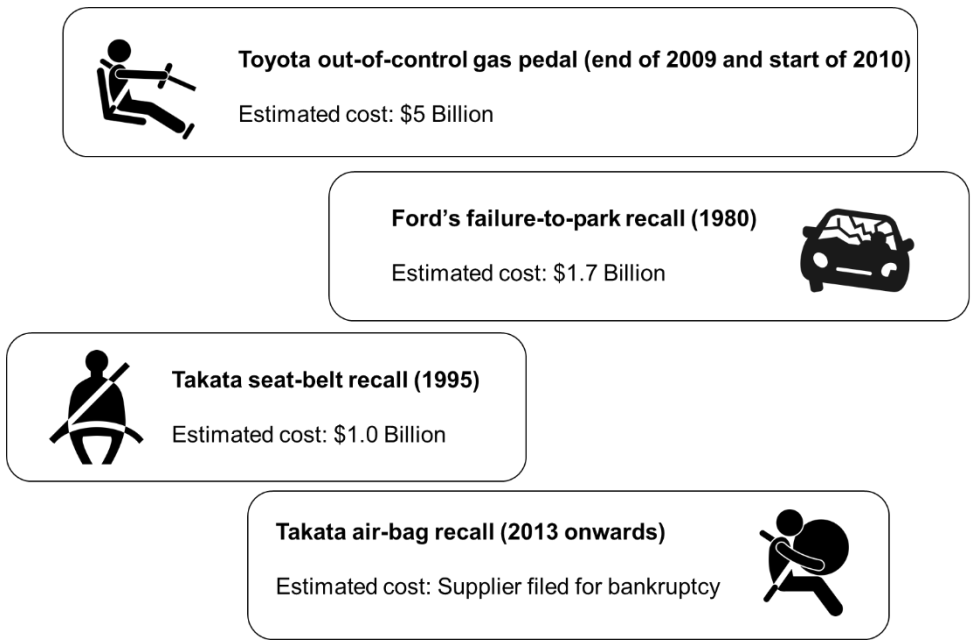

**FIGURE 9.** Example of data driven product development [79].

Apart from vehicle control, end-to-end learning has also been used for other functions such driver distraction recognition [61]. In [61], authors used a pre-trained CNN (VGG19). Despite the challenging aspects considered in the dataset in terms of different illumination conditions, camera positions and variations in driver's ethnicity, and genders, the proposed network gave a best test accuracy of 95% and an average accuracy of 80% per class.

### D. DATA DRIVEN PRODUCT DEVELOPMENT

AI is expected to play a key role in data-driven product development. Over the years, there have been numerous recalls that have cost carmakers billions of dollars (Figure 9). Understanding the potential threat and developing effective ways to prevent, resolve, and eliminate recalls is going to be critical. Carmakers are investing heavily in technology that will help them track the performance of their products from cradle to grave, and thus better anticipate and mitigate the risk of massive recalls.

## IV. DISCUSSION

Ng [62], Co-founder of Coursera, and Adjunct Professor of Computer Science at Stanford University, reinforces the importance of the data, the fuel of DL models. He states: "It's not who has the best algorithm that wins, it's who has the most data". Traditionally, data was in the hands of a select few. Existing corporations could leverage historical datasets which gave them a competitive advantage over new enterprises or individuals. This is likely to change in the coming times as we see a plethora of open source datasets and associated tools becoming available. Table V summarizes some of the largest datasets that have been available for autonomous vehicle (AV) applications. This democratization of data is empowering researchers to conceptualize novel uses cases and make DL a mainstream enterprise technology.

Low cost hardware advancements and connection solutions will be crucial for the commercialization and uptake of this technology at a faster pace. We are starting to see an entirely new breed of computing architectures, tailored for real-time implementation of DL applications [63]. As researchers and developers strive to commercialize some of the use cases, especially in the context of vehicle motion control, a key consideration would be ensuring compliance with standards, such as ISO 26262 [64], that regulates functional safety of road vehicles. These standards describe actions to ensure reduction of risks that are typically caused by malfunctioning hardware and/or software components. Adequacy of DL models from the perspective of safety certification remains controversial [65]. Without knowledge of the model's internals, current testing paradigms are not able to find different corner-cases and erroneous behaviors under different realistic driving conditions [71]. While decisions made by rule-based software can be traced back to the last if and else, the same can't be said about DL algorithms. Hence, the development of testing and verification frameworks for DL models is drawing a lot of attention from researchers [72], [73] and we expect this to remain an active area of research.

TABLE V
OPEN SOURCE DATABASES AND DATA VISUALIZATION TOOLS FOR AUTONOMOUS VEHICLES

| Name | Dataset Owner | Description | Link | Publicly Available Since | Reference |
|---|---|---|---|---|---|
| Berkeley DeepDrive BDD100k | UC Berkeley | Large-scale driving dataset | https://bdd-data.berkeley.edu/ | May 2018 | [66] |
| nuScenes | nuTonomy | Large-scale driving dataset | https://www.nuscenes.org/ | Sep 2018 | [67] |
| Baidu Apolloscapes | Baidu | Large-scale driving dataset | http://apolloscape.auto/ | March 2018 | [68] |
| Autonomous Visualization System | Uber | Open standard for autonomous vehicle visualization | https://avs.auto/#/ | Feb 2019 | [69] |

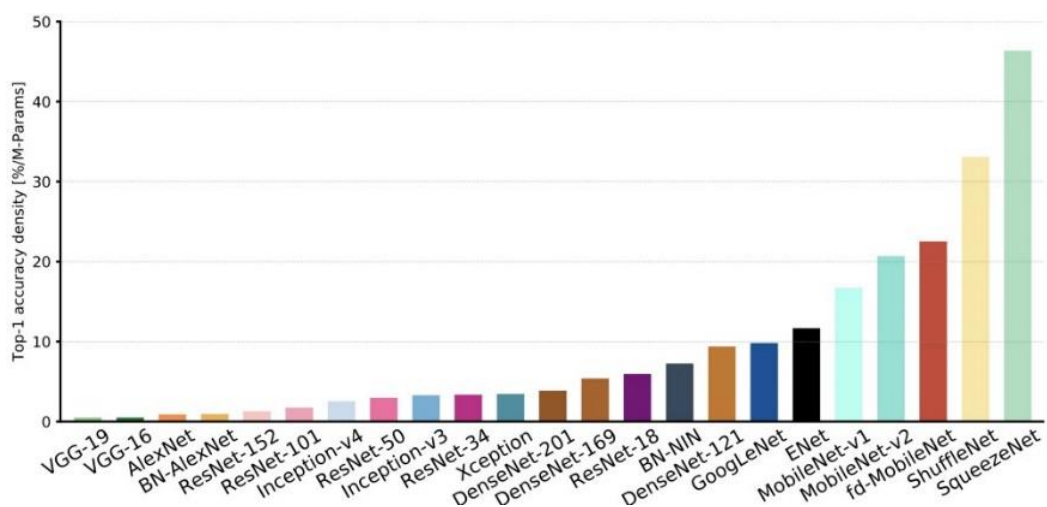

FIGURE 10. Accuracy per parameter vs. network [70] for commonly used CNN models.

DL models are known to learn useful representations and features automatically, thus obviating the need to hand craft the feature vector. However, one has to exercise sufficient caution since some of these models might lack interpretability, which negatively impacts the trust users have in the decisions made by these systems. When accuracy outpaces interpretability, human trust suffers, affecting the adoption rate.

There are on-going attempts to address interpretability in, e.g., CNNs. Visualization techniques like gradient class activation mapping, among other methods, shed light on the models' workings and provide some sanity checks. This is an active area of development as people demand more accountability from DL.

Recent studies [70] have shown how some of the popular CNN networks are highly inefficient in utilizing their full learning power. In Figure 10, we clearly see that, although VGG has a better accuracy than AlexNet, its information density is worse. Information density (accuracy per parameters) is an efficiency metric that highlight that capacity of a specific architecture to better utilize its parametric space [70].

TABLE VI
TYPICAL POWER REQUIREMENTS [76]

| Application | Individual Sensors/Wearables 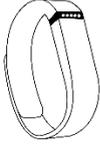 | Level 1 - 3 AV 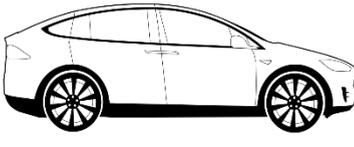 | Level 4 - 5 AV 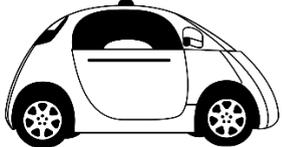 |
|---|---|---|---|
| Power requirements | 500mW – 5 Watts | 10 – 30 Watts | 100's Watts |

Models like VGG and AlexNet are clearly oversized, and do not take fully advantage of their potential learning ability. On the other hand, SqueezeNet achieves AlexNet-level accuracy on the ImageNet dataset with 50x fewer parameters [74]. Some of the techniques employed for developing smaller neural networks include [75]:
- Replacing fully-connected layers with convolutions
- Kernel reduction: Reduction the height and width of filters
- Channel reduction: Reduction of number of filters

Another important consideration for the real-time deployment of these models would be their energy efficiency (Table VI).

Current state-of-the-art DL models used in AVs for object detection require 200 W+ of GPU computing [77]. Design and engineering of energy and memory efficient DL models is expected to be a key area of research [78].

## V. CONCLUSION

The automobile as we know it today is being pulled apart, reimagined and rebuilt. DL is paving the way for the delivery of unprecedented automotive innovations that will disrupt the status quo and deliver an enriched user experience. This paper provides a comprehensive review of relevant works about DL methods applied in the context of automotive related use cases. The field of DL has matured a lot in the last decade, and changed a lot in the last few years. New architectures scaled to be larger/deeper, take advantage of large number of datasets and parallel computing power. We first provide an overview of the key DL architectures being employed for automotive relevant use cases. Supervised DL methods, namely, CNNs and RNNs, that have proven their efficacy in other domains are also the natural choice for researchers in the automotive domain. As an outcome of this survey, novel uses cases have been identified and bucketed into three categories: (1) virtual sensing for vehicle dynamics, (2) vehicle inspection/health monitoring and (3) automated driving. Furthermore, for each of these categories, a systematic review has been conducted to report key findings and also highlight the shortcomings of pertinent reports and papers. Finally, the authors discuss some important aspects such as compute power requirements, model transparency and interpretability, model compliance with vehicle safety standards, all of which are expected to appreciably impact the adoption rate of DL in the automotive industry.

## AUTHOR BIOGRAPHIES

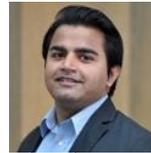

KANWAR BHARAT SINGH is currently a Principal Engineer at the Goodyear Tire and Rubber Co. He received his M.S. degree in Mechanical Engineering from Virginia Tech in 2012 and has since been working at the Goodyear Innovation Centers in Akron, Ohio and Luxembourg. He holds several US patents and has authored numerous peer reviewed technical papers in reputed journals. He is serving as an Associate Editor for the SAE International Journal of Passenger Cars. His current research interests include intelligent tires, vehicle state estimation and machine learning for automotive applications.

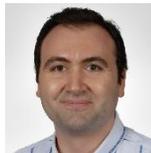

MUSTAFA ALI ARAT is currently a Staff Engineer at the Goodyear Tire and Rubber Co. His research revolves around the fields of transportation systems, vehicle dynamics and controls, signal processing and machine learning methods. He was awarded his Ph.D degree. in 2013 from Virginia Tech and has held positions both in academia and industry. He has authored numerous peer reviewed technical publications and is an active member of ASME and SAE.